# Nowcasting the Financial Time Series with Streaming Data Analytics under Apache Spark

Mohammad Arafat Ali Khan, Chandra Bhushan, Vadlamani Ravi*,

Vangala Sarveswara Rao and Shiva Shankar Orsu

[1]Centre of Excellence in Analytics,
Institute for Development and Research in Banking Technology,
Castle Hills Road No. 1, Masab Tank, Hyderabad-500057, India

arafathkhan7852@gmail.com; cbhushan051@gmail.com; vravi@idrbta.c.in; sarveswararao.cs@gmail.com; shivaorsu96@gmail.com

**Abstract.** This paper proposes nowcasting of high-frequency financial datasets in real-time with a 5-minute interval using the streaming analytics feature of Apache Spark. The proposed 2-stage method consists of modelling chaos in the first stage and then using a sliding window approach for training with machine learning algorithms namely Lasso Regression (LR), Ridge Regression (RR), Generalised Linear Model (GLM), Gradient Boosting Tree (GBT) and Random Forest (RF) available in the MLLib of Apache Spark in the second stage. For testing the effectiveness of the proposed methodology, 3 different datasets, of which two are stock markets namely National Stock Exchange (NSE) & Bombay Stock Exchange (BSE), and finally One Bitcoin-INR conversion dataset. For evaluating the proposed methodology, we used metrics such as Symmetric Mean Absolute Percentage Error (SMAPE), Directional Symmetry (DS) and Theil's U Coefficient. We tested the significance of each pair of model's using the Diebold Mariano (DM) test.

## 1. Introduction

As we move rapidly towards digital technology in service industries, the amount and the speed with which the data is generated in day-to-day activities is truly spectacular, signalling the presence of the volume and the velocity dimensions of Big Data. To make meaningful decisions from this kind of data, we need to make use of tools which can effectively handle huge amounts of data coming at a fast pace. Both batch processing the historical stored data and real-time stream processing of data are the hallmarks of the Big Data technologies. If the data is generated in real-time, then we can make use of big data analytics to derive the insights and take more informed timely decisions based on it. Spark Streaming is capable of effectively handling real-time data. It is an extension of the Core Spark API which enables high throughput, fault-tolerant and highly scalable processing. Spark streaming receives live data streams and divides the data into batches and then it processes them using the Spark engine to generate the stream of results in batches.

---

* Corresponding Author; Tel.: +914023294310; Fax: +914023534551; E-mail: rav_padma@yahoo.com

Nowcasting using financial data has become a recent trend due to advances in machine learning and big data and the advent of new distributed and in-memory computing platforms. The term nowcasting is a portmanteau word for now and forecasting and has been used for a long time in meteorology and recently also in economics. Nowcasting refers to the prediction of the present, the very near future and the very recent past. The principle behind nowcasting is making use of real-time data streams at high frequencies to obtain an early estimate of the market.

This paper proposes nowcasting of Stock Prices and bitcoin-INR exchange rates under big data analytics paradigm. The main focus is on using the real-time data generated from these financial time series to forecast the very near future in the series using the machine learning algorithms. To the best of our knowledge, no one proposed to use big data analytics for nowcasting yet.

The rest of the paper is organised as follows: section 2 presents the motivation to the work; section 3 presents the contribution made in this work. Section 4 presents the state-of-the-art in Financial Forecasting. The proposed methodology and experimental setup are described in section 5. Section 6 presents the information about the Dataset and evaluation metrics. The Results and Discussion are presented in Section 7. The paper is concluded with some future directions of the work in section 8.

## 2. Motivation

In the literature, to the best of our knowledge, not a single study was reported to real-time forecasting (a.k.a nowcasting) of financial time series under big data paradigm. No study dealt with two distinct financial time series data, namely, stock market price, and cryptocurrency price in a single paper.

## 3. Contribution

We streamed real-time high frequency data at the intervals of 5 minutes from 3 different data sources and performed nowcasting experiments in a parallel processing environment in a YARN cluster. Real-time sliding window approach was implemented in our experiment to nowcast. We have adapted 5 different methods available in Machine learning library of Spark. The same study can be implemented in nowcasting the macroeconomics time series data like consumer price index (CPI), if such a data is available and can be streamed at a high frequency.

## 4. Literature Survey

Bańbura et al. [1], and subsequently 2013 [2] defined the nowcasting as predicting the present, a very near future, where they nowcast the GDP of fourth-quarter using the multivariate dynamic model in the European region, and United State of America (USA) respectively. Giannone et al. [3] illustrated USA

GDP current quarter nowcast using large factor models (FM) i.e. a large number of indicators around 200 which lead to the "curse of dimensionality" and later sorted by constructing principal components of the indicators. The parameters for the forecasting model are estimated using Ordinary Least Squares (OLS) regression. But due to staggered-data, this nowcast is "pseudo-real-time". Richardson et al. [4] performed New Zealand GDP quarterly nowcast using popular machine learning techniques namely Lasso, Ridge, Elastic Regression, and Least-squares boosting on historical data provided by Reserve Bank of New Zealand. Each model is evaluated using Symmetric Mean Absolute Percentage Error (SMAPE) and Mean Absolute Deviation (MAD). They reported that the ML techniques used by them outperformed the traditional statistical methods.

Ilamchezhian et al. [5] published a review on spark framework in 2017, where they discussed the features of spark, and its advantages and disadvantages. They also brought up the problem of using the Hadoop Distributed File System (HDFS) i.e. speed of transferring data from HDFS for processing is higher, which is solved by the Spark RDD (Resilient Distributed Dataset). The RDD is an immutable tuple of partitioned data, distributed across the cluster. Spark framework provides the Streaming and MLLib libraries which we have exhausted at full in our experiments. As spark processing overcomes the speed issue of HDFS, we setup spark cluster over HDFS in Yarn mode. This setup framework gives us the freedom to stream and perform as many datasets as we require.

Modugno [6] used the Dynamic Factor Model (DFM) to nowcast inflation by including the high-frequency data and raw material prices in the US. Bok et al. [7] also illustrated the DFM is parsimonious for high dimensional data. In another nowcasting experiment, Loermann et al. [8] in 2019, performed nowcasting of US GDP using artificial neural network (ANN) with DFM. They collected data from FRED-MD from 1999 to 2018 to predict in quasi-real-time. In 2016, Jayanthi et al. [9] described different tool to perform big data analytics in real-time. They mentioned so many tools, cloud platforms like Hadoop, Apache-Spark, Flume, Sqoop, AWS, Amazon Kinesis and so on. They also illustrated the comparative study of various resources and provided architecture to perform big data analytics.

Ranjitha, B.P. [10] in the survey paper explained how real-time analytics help in monitoring data generated from sensors. The paper describes different sources of real-time data generated from different sensors and how to handle a large amount of data by using Big Data Framework. The author mentioned different processing techniques, Complex Event Processing (CEP) for the streaming data and how they monitor with the help of visualisation. Girija et al. [11] stock price prediction is based on two types of prediction analysis: technical and fundamental. Along with Historical Stock market data, the social media data was used here for the stock price prediction. They implemented on news articles, tweets of the company's and historical stock data from Yahoo finance, whenever real-time data is generated it is streamed using flume and stored in the HDFS. Logistic Regression is modelled to predict the next day trend. In [12] the movement of the opening of the stock price of a company is predicted using sentiment of the user, for data ingestion they have used Twitter streaming API worked upon the spark streaming then dumped the data into HDFS using Apache Flume. In the paper, Lambda architecture was used for entire processing and word embedding extraction of a sentiment RNN is deployment, analysis is done on Spark's architecture using model Naïve Bayes. The experiment is executed on the data of Google, Apple and Microsoft. In 2016, Menon et al. [13] performed a similar experiment where they used spark streaming using Scala as big data architecture to complete this task. They used multiple indices from the NSE stock market to predict the minute-wise indices price parallel. They took 50 min window size, 25 minutes sliding interval and run 20 experiments on spark. They applied ARMA to predict the prices of the indices. Umadevi et al. (2018) [14] similar to [12] the datasets of Google, Apple and Microsoft are used for the experiments. They trained the ARIMA model on historical data close parameter of all firms and 25 values/scores are forecasted.

In another work, WU et al. [15] described the chaotic modelling using lag and embedding dimension where they calculated the lag and embedding dimension using Fast Orthogonal programming and Wavelet Network. After that, they applied the Wavelet Neural network for forecasting the spot market price of electricity. In 2017, Selvin et al. [16] worked on the dataset of NSE listed companies Infosys, TCS and CIPLA with minute wise stock price with a sliding window approach of size 100 minutes with 90 minutes as training and 10 minutes for prediction. They Implemented Deep learning architectures CNN, RNN and LSTM for these experiments and made a comparison with linear model ARIMA in terms of RMSE score. Similar Modelling with LSTM and ARIMA was Implemented by Raju and Tarif (2020) [17] in predicting real-time Bitcoin price using the merged dataset of Bitcoin historical and real-time data collected from different public API and the other dataset is tweets related to the Bitcoin Collected using Twitter API on which sentiment analysis is performed.

In this work, we have tested out the proposed methodology on three varieties of datasets namely stock market, foreign exchange, and cryptocurrency conversion. We have setup a Yarn cluster to perform parallel operations on all the datasets at the same time. WU et al. [15] applied chaotic modelling before applying any forecasting algorithm, but in this work, we used a different procedure for chaotic modelling. Menon et al. [13] and Selvin et al. [16] applied sliding window approach using Scala on spark cluster but we applied the sliding window approach using Pyspark on Spark cluster in Yarn mode with different window size and sliding interval. Besides this, the main focus of this study lies in training the models in real-time for nowcasting.

## 5. Proposed Methodology

We have collected stock market historical data (300 data points with the frequency of 5 minutes) from Yahoo finance API and the data is stored on HDFS in a multi-node YARN cluster. Our methodology has two stages: 1. Phase space Reconstruction and 2. Nowcasting using ML models. In stage 1, by using

Lyapunov exponent we found chaos in our datasets, so we performed chaotic modelling by calculating the lag/time delay and embedding dimension using PACF function available in python package and Cao's algorithm available in CRAN package of R. This process needs to be repeated over a daily basis before starting the experiment as lag and embedding dimension is observed to change with time.

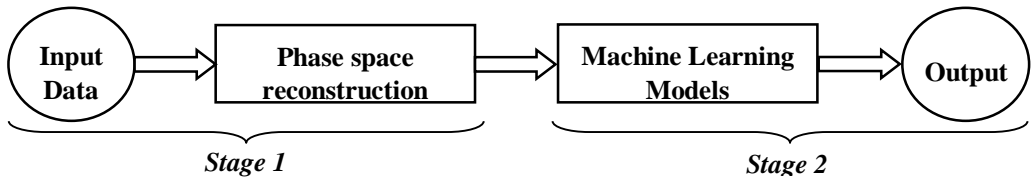

*Figure 1* *Stages of Proposed Methodology*

After calculating the time lag and embedding dimension, we applied TakensEmbedding to reconstruct datasets from time series to supervised form (Multi-Input Single-Output), which can be modelled by any forecasting algorithm. In stage 2 for nowcasting we have implemented 5 different Machine learning models, namely, Lasso Regression, Ridge Regression, Random Forest, Gradient Boosting Tree(GBT), and Generalised Linear Model(GLM) available in Spark Ml library. We found the best models for each dataset by tuning each model using the list of hyperparameters mentioned in Table 10. For nowcasting, we made use of the below equation (1).

$$\hat{y}_{t+1} = f(y_{t-(emb-2)*lag}, y_{t-(emb-3)*lag}, y_{t-(emb-4)*lag}, \dots y_t) \quad (1)$$

where $\hat{y}_{t+1}$ is forecasted value, f is the function takes actual values to calculate $\hat{y}_{t+1}$, $y_t$ is the actual value at time t, *emb* is embedding dimension, and *lag* is the time delay.

In other words, let's say we have a lag of 1 and embedding dimension of 5 and have 300 data to train the model. So after phase space reconstruction we got a matrix of 60 rows and 5 columns shape. Then we consider 1st to 4th columns as Independent features and 5th columnas target. Consider forecasting the 301st data point and for that, we will take the previous $y_{297}$, $y_{298}$, $y_{299}$, $y_{300}$ values using the equation (1). As $\hat{y}_{t+1}$ is forecasted, we will compare the forecast with the actual value and calculate the squared error. The algorithms will re-train to best fit the available data if the forecasted Squared Error varies by more than 5% when compared to training MSE otherwise then there will be no need of training and the model will continue to forecast next value in the series.

After forecasting, we consecutively stream the actual value from the source API using Spark Streaming framework, and then we appended the actual value to the training data, if re-training is needed when the condition of forecasted squared error varies more than 5% of training MSE. In this case, the training data consists of previous training data + newly appended data points. We have used a sliding window of size 300, which slides over 300 data points and forecasts the 301ˢᵗ value, in the next step it leaves the first point behind and appends the new data point to the training set for every training period and re-train the model. For example, assume if we have a training set of 10 data points,

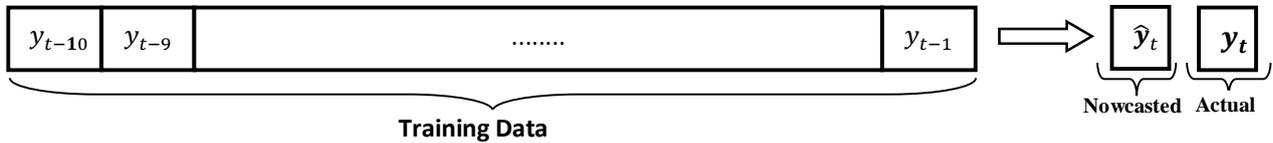

Using the trained model we forecast the $\hat{y}_t$, we also have $y_t$ (actual value at that time stamp streamed using API). Here we check the squared error of actual and forecasted value and compare with the training MSE and if it varies more than 5%, then that denotes our model had forecasted a bad value and we need to tune the hyperparameters and retrain the model. Here we need to retrain the model by adding the $y_t$ to the training set leaving the first data point in the training set nothing but a sliding window of 10 in this example.

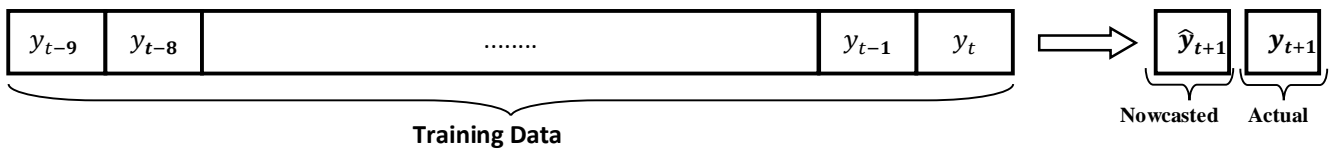

As we are nowcasting at high frequency of time intervals we are not making the bad value forecasted ($\hat{y}_t$) correct/more accurate, rather we are re-training our model to improve the forecasting in the next time stamp i.e., $y_{(t+1)}$

The forecasting is done for the entire day (i.e. till the market is closed) with the same frequency of 5 minutes. For the next day, we take the previous 4 days' historical data and find the lag and embedding of the data to convert the series into MISO problem followed by nowcasting using algorithms. In this way, we have conducted experiments consequently for 5 days.

Since we applied 5 models simultaneously on the same dataset, we needed to decide the best one. After the entire day of the forecast, we use three metrics to evaluate the models, i.e. SMAPE, Directional Symmetry, and Theil's U Statistics. We also applied the Diebold-Mariano test to check the significance between models.

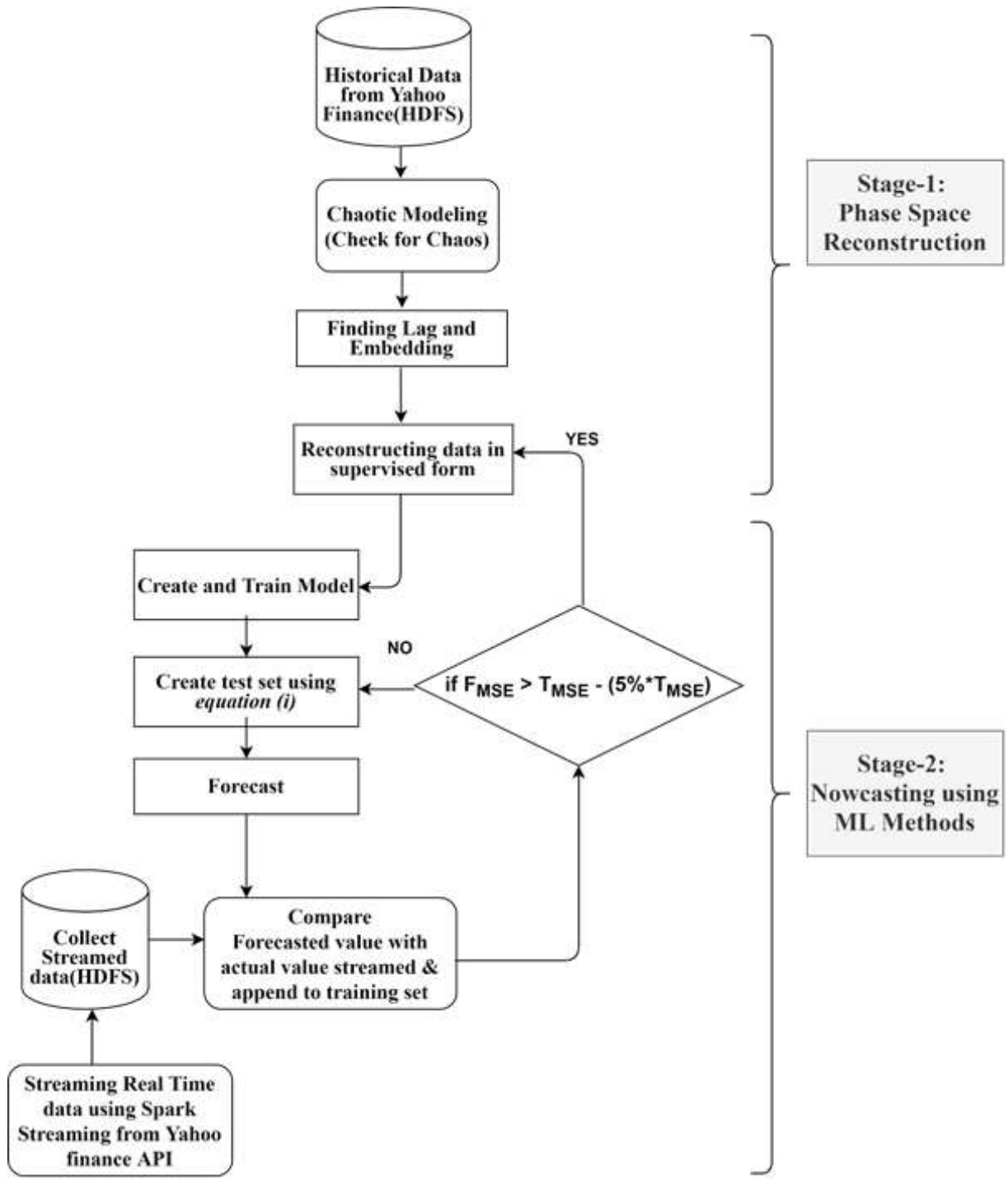

**Figure 2.** The architecture of proposed methodology.

## 5.1 Introduction to Chaos Theory

The theory of Chaos was proposed by Poincare in the late 1800s and later in 1963, the work was extended by Lorenz [18] to deal with complex nonlinear systems [19]. A chaotic system is deterministic, dynamic and evolves from the initial conditions and can be described in state space by the trajectories. The state space is represented by phase space, as the governing equations for the chaotic system are not known. The phase space can be rebuild using the original series using delay time and embedding dimension [20] as shown in Eq. (2).

$$Y_j = (X_j, X_{j+\tau}, X_{(j+2\tau)}, \ldots \ldots, X_{j+(m-1)\tau}) \qquad (2)$$

Whereas Y is target variable, X is independent feature, $\tau$ is the time delay and *m* is the embedding dimension. After rebuilding the phase space, the univariate time series problem can be converted into a multi-input single-output (MISO) problem.

### *5.1.1. Rosenstein's Method*

Rosenstein's algorithm [21] estimates the largest Lyapunov exponent ($\lambda$) [22] from the given time series. If $\lambda \geq 0$ then it contains chaos; otherwise the chaos is not present in the given time series.

### *5.1.2. Cao's Method*

To find the embedding dimension for a given time series Cao's proposed [23] a method. Let $X = (X_1, X_2, X_3, X_4, \ldots, X_N)$. In the phase space, the time series can be reconstructed as time delay vectors as in Eq. (3).

$$Y_i = (X_i, X_{(i+\tau)}, X_{(i+2\tau)}, \ldots \ldots, X_{i(m-1)\tau}) \qquad (3)$$

Where $Y_i$ is target variable, $X_i$ is independent features, m is embedding dimension and $\tau$ is lag or time delay.

### *5.1.3. Taken's Embedding*

After calculating lag and embedding dimension of datasets we need to reconstruct the dataset to transform the time-series data ($x_1, x_2, \ldots, x_n$) to supervised form using the equation $x_{n-(m-1)*\tau}$, where m is the embedding dimension and $\tau$ is the time delay.

## **5.2. Brief Overview of Techniques employed**

### *5.2.1. Lasso Regression*

Lasso regression is a type of linear regression algorithm, in which a regularisation parameter was introduced by Tishbirani. (1996) [24]. In this algorithm, the regularisation penalty minimizes the residual sum of errors to the sum of the absolute values of the coefficients for each feature and do feature selection and useful in not overfitting to the data. It is also called an L1 regularisation.

$$Minimize \left( \sum_{i=1}^{n}(y_i - \sum_{j} x_{ij}\beta_j)^2 + \lambda \sum_{j=1}^{p}|\beta_j| \right)$$

Where $y_i$ is target variable $x_{ij}$ independent variable of $j^{th}$ row and $i^{th}$ column, β is learnable parameter and $\lambda$ is hyperparameter for penalising the learnable parameter.

*5.2.2. Ridge Regression*

Ridge Regression is similar to Lasso Regression. This performs L2 regularization in which the penalty equivalent is added to the square of the magnitude of the coefficients.

$$Minimize \left( \sum_{i=1}^{n}(y_i - \sum_{j} x_{ij}\beta_j)^2 + \lambda \sum_{j=1}^{p} \beta_j^2 \right)$$

Where $y_i$ is target variable $x_{ij}$ independent variable of $j^{th}$ row and $i^{th}$ column, $\beta$ is learnable parameter and $\lambda$ is hyperparameter for penalising the learnable parameter.

*5.2.3. Random Forest (RF)*

Random forest is ensemble learning proposed by Tin (1995) [25]. Random Forest is used for both classification and Regression tasks. It operates by constructing multiple decision trees during training and takes the average of all the results from all individual trees.

*5.2.4. Gradient Boosting Tree (GBT)*

Gradient Boosting is an ensemble of weak prediction models i.e., decision trees. It is proposed by Friedman, (2001) [26]. Boosting is an ensemble technique used to develop a solid model using numerous weak models by including models ahead of one another. For example, the slip-ups or mistakes made by past models are rectified by the next model until the errors are accurately anticipated.

*5.2.5. Generalised Linear Model (GLM)*

The generalised linear model is a generalized version of various linear models, formulated by Nelder and Wedderburn [27]. They proposed an iteratively reweighted least-squares strategy for most extreme probability assessment of model boundaries. The least-square strategy is the most commonly used method in other statistical algorithms. The other methods are Bernoulli, Log-linear, and Poisson distribution are present.

## 5.3. Experimental Setup

A Spark Cluster is configured for computations with 3 nodes as slaves and 1 master upon Hadoop cluster; here we have configured Hadoop for the use of its Hadoop Distributed File System (HDFS) for storing the data in a distributed environment. To submit multiple Job applications on a cluster we have configured YARN upon Hadoop which shares the available resources to the Jobs submitted. For these experiments, a total of 6 different Clusters are used with the same configuration. Each machine has 32 GB RAM with 224 GB SSD to process fast data transfer between disk and RAM. The application can be allocated a minimum of 1 core and a maximum of 6 cores in a container. This sums up to have 90 GB of memory in a single cluster.

### *5.3.1. Streaming Methodology*

Spark Streaming is a core Spark API allows stream processing of live data streams. The data can be ingested from multiple sources like Kafka, Kinesis or TCP sockets. The processed data can be dumped into a local file system or Distributed file system.

Spark Streaming provides the high level of abstraction called DStreams (discretized stream) which is a collection of sequence of RDDs to stream the continuous data streams from any source. In fact, we can apply Spark's machine learning algorithms on the collection of DStreams.

In our experiments we have used the Streaming framework to stream the continuous data from Yahoo finance API and dumped the data into Hadoop Distributed File System (HDFS) and then we streamed the data for different pyspark applications from HDFS.

We streamed the data from source with the help of readStream, which is the function available in Spark Streaming which supports Structured Streaming with the help of DataStreamReader Interface. DataStreamReader is an Interface that reads the source Data and converts it into Streaming Data which is Dstreams. This reads the Data and generates continuous DStreams, and further DStreams are captured and Analyzed.

## 6. Dataset description and evaluation metrics

We have analysed three different datasets namely Bombay Stock Exchange (BSE), National Stock Exchange (NSE), and Bitcoin to Indian Rupee (BTC-INR) conversion. We have continuously streamed the real-time data with a frequency of 5 minutes from these markets using an API provided by yahoo finance. We have used the Closed value from all of the markets. As for BSE, NSE and BTC INR operate from 9:30 a.m. to 3:30 p.m. five days in a week (Indian Market time), we have conducted experiments with their respective operating time on 28$^{th}$, 29$^{th}$ 30$^{th}$ of September and 1$^{st}$ and 5$^{th}$ of October in IST Zone. Whereas the FOREX exchanges operate round the clock, but we have conducted the experiments on these datasets as per the Indian market time.

### 6.1 Evaluation metrics

We tested the models using the three main evaluation metrics that we come across in the literature. They are Symmetric Mean Absolute Percentage Error (SMAPE), Directional Statistics (DS) and Theil's U Coefficient.

### 6.1.1. Symmetric Mean Absolute Percentage Error (SMAPE)

This metric is not biased towards the scale of the datasets like Root Mean Squared Error. SMAPE gives us the error percentage as shown below.

$$SMAPE = \frac{100}{n}\sum_{i=1}^{n}\frac{|\hat{y}_i - y_i|}{(|\hat{y}_i| - |\hat{y}_i|)/2}$$

Where $\hat{y}_i$ is the forecasted value, $y_i$ is the actual value at that time stamp and n is the number of data points forecasted.

### 6.1.2. Directional Symmetry

Directional Symmetry evaluates the performance of the model by predicting the direction of change, whether positive/negative. Its equation form is shown below:

$$DS(t,\hat{t}) = \frac{100}{n-1}\sum_{i=2}^{n}d_i$$

$$\text{where } d_i = \begin{cases} 1, & if(y_i - y_{i-1})(\hat{y}_i - \hat{y}_{i-1}) > 0 \\ 0, & Otherwise \end{cases}$$

Where $\hat{y}_i$ is the forecasted value, $y_i$ is the actual value at that time stamp and n is the number of data points forecasted.

### 6.1.3. Theil's U Coefficient

Theil's U coefficient is a relative accuracy measure that gives more weightage to the large errors that helps in choosing models with less error. If the value of U is greater than 1 then the naïve guess is better than forecast and the closer of U value to zero, the better the forecast method.

$$Theils\ U\ Coefficient = \sqrt[2]{\frac{\sum_{t=1}^{n-1}(\hat{y}_{t+1} - y_{t+1})^2}{\sum_{t=1}^{n-1}\left(\frac{y_{t+1}-y_t}{y_t}\right)^2}}$$

Where $\hat{y}_{t+1}$ is the forecasted value at t+1 time stamp, $y_{t+1}$ is the actual value at t+1 time stamp, and n is the number of data points forecasted.

### 6.1.4. Diebold Mariano Test

DM test is proposed by Diebold and Mariano (1995) [28]. It is a statistical test to check the equality of forecast models. It determines whether forecasts of two or more models are significantly different or not.

## 7. Results and Discussion

After nowcasting, we have evaluated our models by metrics namely, SMAPE, Directional Symmetry, and Theil's U Coefficient. To check the significance between model forecasting we have included Diebold's Mariano Test. Table 1 presents the combined 5 days forecast results of Lasso, Ridge, Random Forest, Gradient Boosting tree and Generalised Linear Regression across all three datasets in terms of SMAPE. In terms of SMAPE, GLM performed better than other algorithms for all the datasets followed by Ridge regression. The ensemble learning capability of Random forest did not seem to give any better results than less complex Ridge or Lasso Regression across in terms of SMAPE and Theil's U Coefficient.

*Table 1: SMAPE on Combined 5 Days results*

| Dataset / Model | Lasso | Ridge | RF | GBT | GLM |
|---|---|---|---|---|---|
| **BSE** | 0.07077 | 0.06758 | 0.12980 | 0.11409 | **0.06754** |
| **NSE** | 0.06642 | **0.06497** | 0.11913 | 0.11025 | **0.06497** |
| **BTC-INR** | 0.05240 | **0.04915** | 0.07918 | 0.06836 | 0.04921 |

When we computed the Directional Symmetry (see Table 2.) for all algorithms across all datasets, GBT is giving better directional correlation to the actual values than all the algorithms experimented. Here the less complex models such as Lasso and Ridge are not able to capture the directional symmetry when compared with more complex models such as RF, GBT, and GLM.

*Table 2: Directional Symmetry on Combined 5 days results*

| Dataset / Model | Lasso | Ridge | RF | GBT | GLM |
|---|---|---|---|---|---|
| **BSE** | 44.87 | 44.59 | 50.41 | **51.24** | 44.59 |
| **NSE** | 45.55 | 48.51 | **52.56** | 49.32 | 48.51 |
| **BTC-INR** | 53.16 | 53.71 | 47.93 | 51.51 | **53.99** |

In terms of Theil's U coefficient values (see Table 3), both GLM and Ridge are outperforming all the models but they showed similar Theil's U Coefficient values because regularization constants of GLM and Ridge are only differed by a small margin in this case. Here Ridge regression even outperformed more complex models such as GBT and RF by a very large margin.

*Table 3: Theil's U Coefficient for Combined 5 days Results*

| Dataset / Model | Lasso | Ridge | RF | GBT | GLM |
|---|---|---|---|---|---|
| **BSE** | 0.700 | **0.693** | 1.261 | 1.319 | **0.693** |
| **NSE** | **0.703** | 0.705 | 1.280 | 1.298 | 0.705 |
| **BTC-INR** | 0.428 | **0.419** | 0.631 | 0.563 | **0.419** |

In terms of BSE data, with respect to all 5 days forecast the Generalised linear model (GLM) performed better in terms of SMAPE which is followed by Ridge Regression. The least performer with BSE data was GBT with a high percent of SMAPE. But, GBT was giving a better directional correlation when compared to other algorithms. Ridge and GLM are the same but better than others when compared in terms of Theil's U coefficient.

In NSE data, the combined SMAPE of all the 5 days forecast of Ridge Regression and GLM are almost the same and performed better than others. While looking into Directional Symmetry of NSE, the Random Forest was good in showing the directional correlation. Lasso Regression Theil's U coefficient is closer to zero when compared to rest algorithms.

For BTC-INR, the same Ridge Regression and GLM having low percent of error in terms of SMAPE. GLM proved to be better in both metrics Directional Symmetry and Theil's U coefficient.

In Diebold Mariano Test, we compared two models at alpha = 0.05 on SMAPE metrics where if the DM test is greater than the p-value then the model is significantly different from each other and the models with less SMAPE value is better one. And if the DM test is lesser than p-value then the models are not significantly different.

In Table 4, we compared all models with each other and found some are significantly different and some are not on the BSE dataset. The models which are significantly different is highlighted with light orange and in that model which is highlighted in BOLD in the below tables had better forecast based on SMAPE value and another insignificant models are highlighted with light green. We found that Ridge-Lasso, GBT-Random Forest and GLM-Ridge forecast are almost the same. Likewise, in Table 5 and 6are the DM test performed on the NSE and BTCINR datasets respectively.

*Table 4: Diebold Mariano for BSE Results.*

| Model_1_Forecasts vs Model_2_Forecasts | DM | P-Value |
|---|---|---|
| Ridge vs Lasso | 0.3499 | 0.7265509 |
| Random Forest vs **Lasso** | 4.1153 | 4.79E-05 |
| Random Forest vs **Ridge** | 4.2901 | 2.30E-05 |
| GBT vs **Lasso** | 1.5652 | 0.1184116 |
| GBT vs **Ridge** | 1.5761 | 0.1158637 |
| **GBT** vs Random Forest | 0.2109 | 0.8330729 |
| GLM vs Lasso | 0.3679 | 0.7131292 |
| **GLM** vs Ridge | 1.6080 | 0.1086908 |
| **GLM** vs Random Forest | 4.2920 | 2.28E-05 |
| **GLM** vs GBT | 1.5766 | 0.1157492 |

*Table 5: Diebold Mariano for NSE Results.*

| Model_1_Forecasts vs Model_2_Forecasts | DM | P-Value |
|---|---|---|
| Ridge vs Lasso | 0.1516 | 0.8795701 |
| Random Forest vs **Lasso** | 3.5442 | 0.0004440 |
| Random Forest vs **Ridge** | 3.5842 | 0.0003832 |
| GBT vs **Lasso** | 2.4581 | 0.0144199 |
| GBT vs **Ridge** | 2.4679 | 0.0140409 |
| GBT vs Random Forest | 0.2655 | 0.7907101 |
| GLM vs Lasso | 0.1512 | 0.8798286 |
| **GLM vs Ridge** | 1.0004 | 0.3177584 |
| **GLM** vs Random Forest | 3.5842 | 0.0003832 |
| **GLM** vs GBT | 2.4679 | 0.0140406 |

*Table 6: Diebold Mariano for BTC_INR Results*

| Model_1_Forecasts vs Model_2_Forecasts | DM | P-Value |
|---|---|---|
| **Ridge** vs Lasso | 0.6860 | 0.4931104 |
| Random Forest vs **Lasso** | 5.0178 | 0.0000008 |
| Random Forest vs **Ridge** | 4.7444 | 3.01E-06 |
| GBT vs **Lasso** | 3.5216 | 0.0004835 |
| GBT vs **Ridge** | 3.4033 | 0.0007399 |
| **GBT** vs Random Forest | 4.3945 | 1.46E-05 |
| **GLM** vs Lasso | 0.6535 | 0.5138093 |
| GLM vs **Ridge** | 1.0616 | 0.2890815 |
| **GLM** vs Random Forest | 4.7343 | 0.0000031 |
| **GLM** vs GBT | 3.3944 | 0.0007634 |

From the Figures 3,4 and 5 we can see that again Ridge and GLM are giving closer forecasts to the actual series values compared to other models experimented in this study. They are continuously performing better even with the fewer data in Day_1 and Day_2. Another detail from the figures is that the random forest and GBT are generating the worst forecasts compared to the remaining models even though both of them are ensemble models.

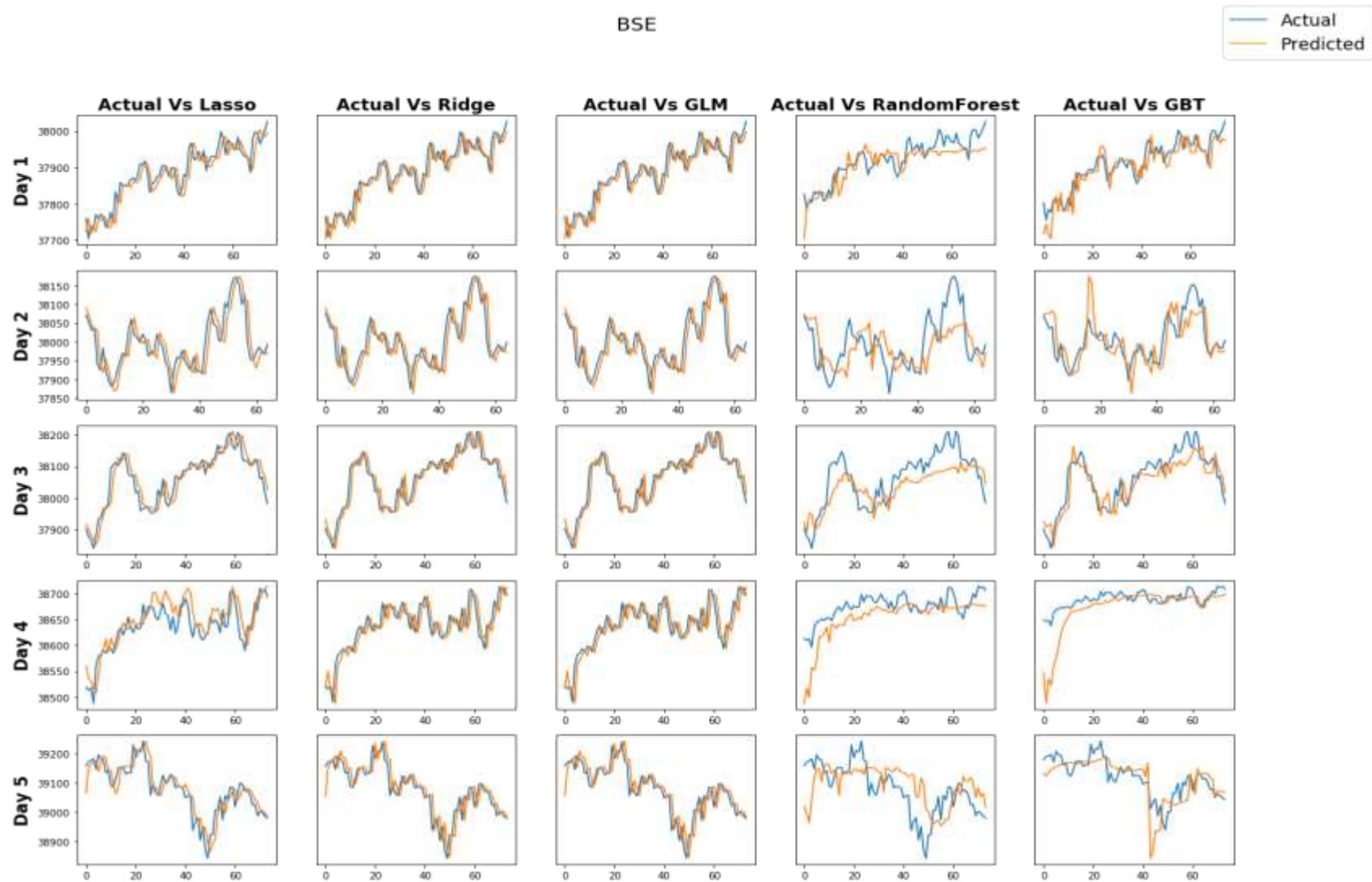

**Figure 3.** Bombay Stock Exchange (BSE) Day-Wise forecasts plots.

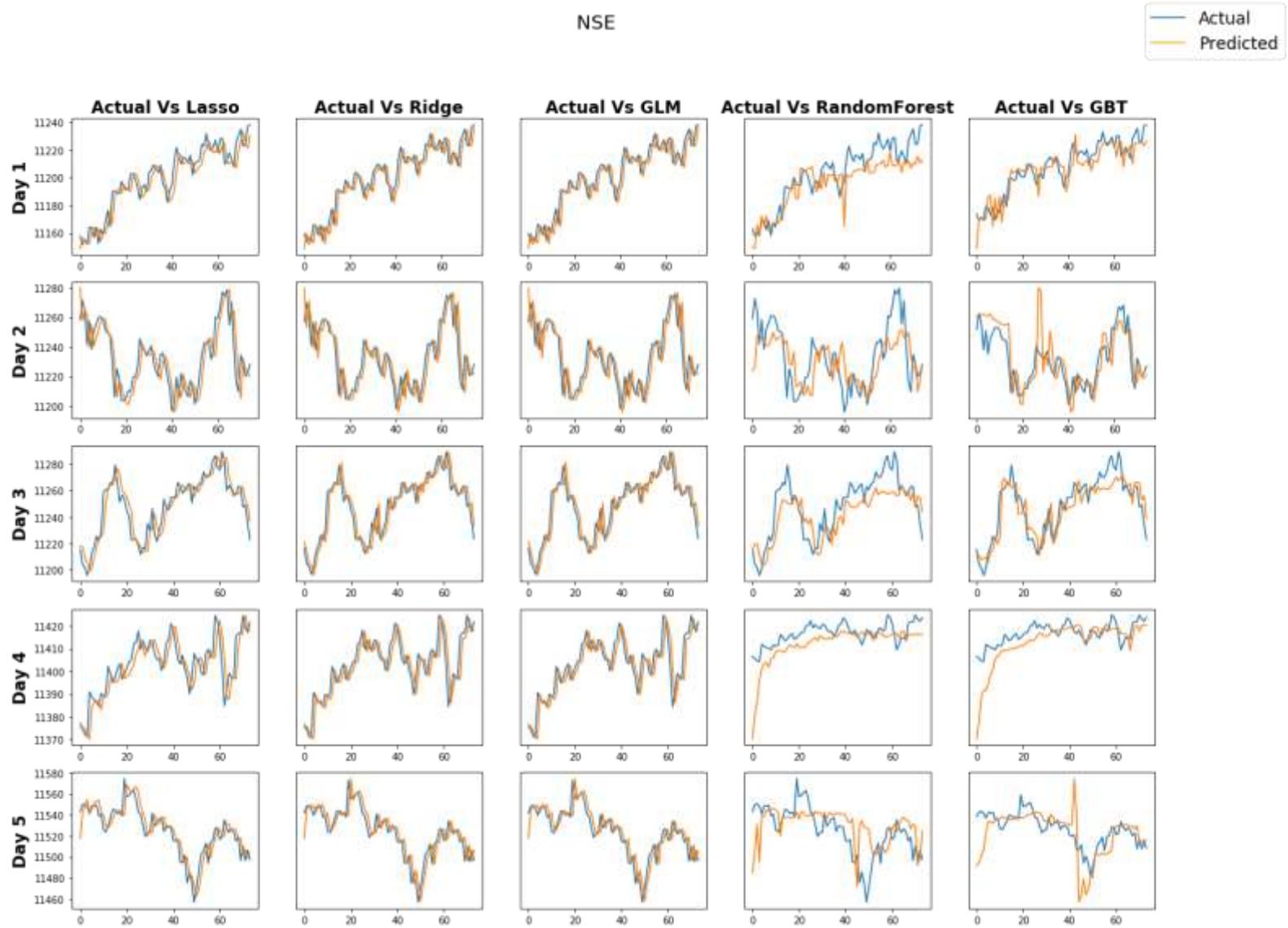

**Figure 4.** National Stock Exchange (NSE) Day-Wise forecasts plots.

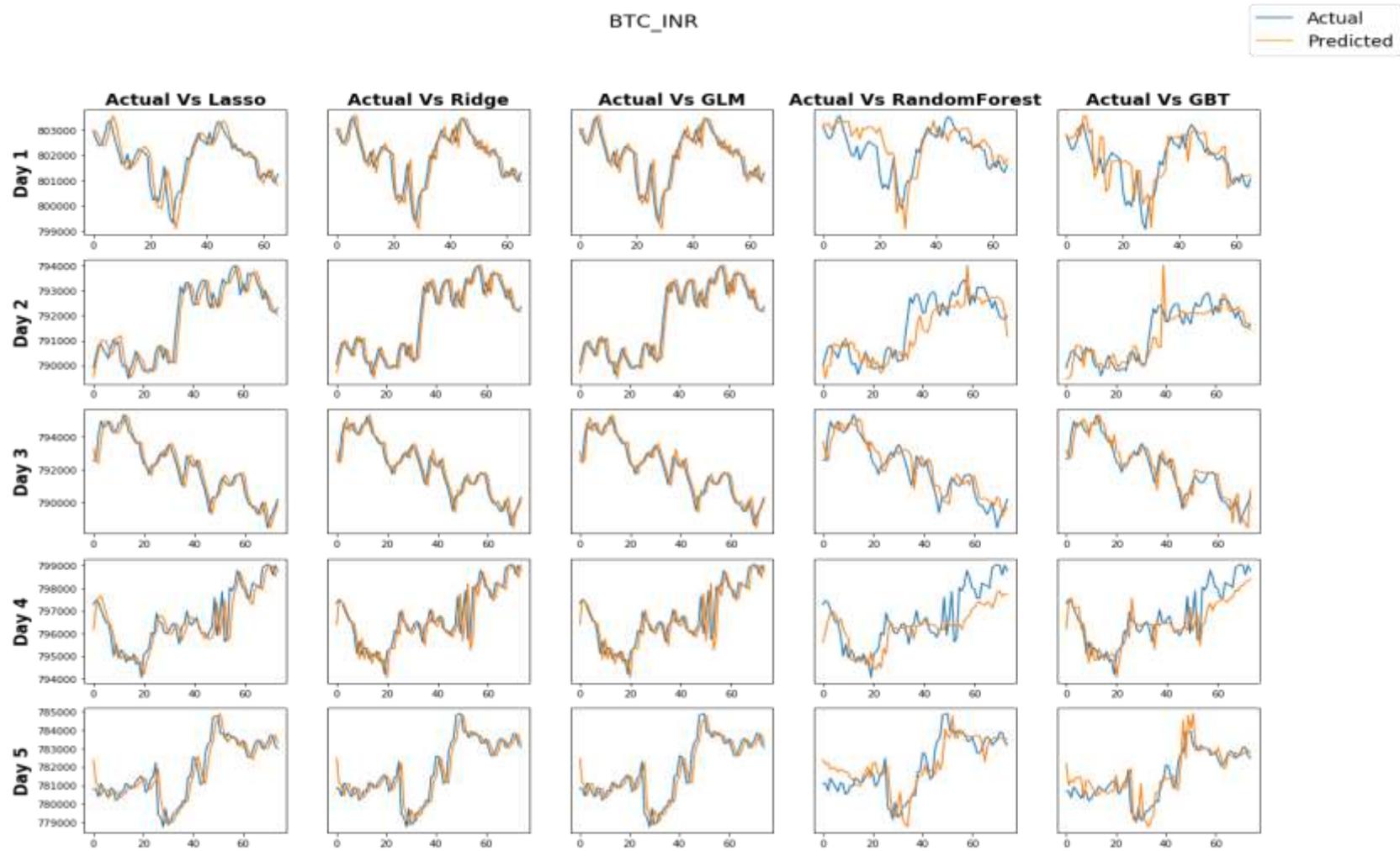

*Figure 5.* BTC-INR Day-Wise forecasts plots.

Figures 9, 10 and 11 depict the Box-plots for all the algorithms on a daily basis using SMAPE values. Using box plots, we can easily identify the model with high variability in results. i.e. model with less Inter Quartile Range (IQR) and less number of outliers is the best model. First we have BSE and then followed by NSE and BTC-INR. Each chart contains 5 different boxplots where each boxplot shows models vs their corresponding SMAPE day wise. We can conclude the lesser the IQR of the boxplot and presence of less no. of outlier will be the better of all. We can see few models having less IQR and no outlier. In most of the dataset Lasso, Ridge, and GLM having less IQR and no outliers whereas RF and GBT have high IQR and lots of outliers. We can observe from the boxplot of $4^{th}$ day forecast of all the algorithms, the IQR is comparatively lesser than the rest of the days. Presence of high number of outlier represents that the model is not able to forecast the price efficiently. So we can conclude that Lasso, Ridge and GLM performed better than other models.

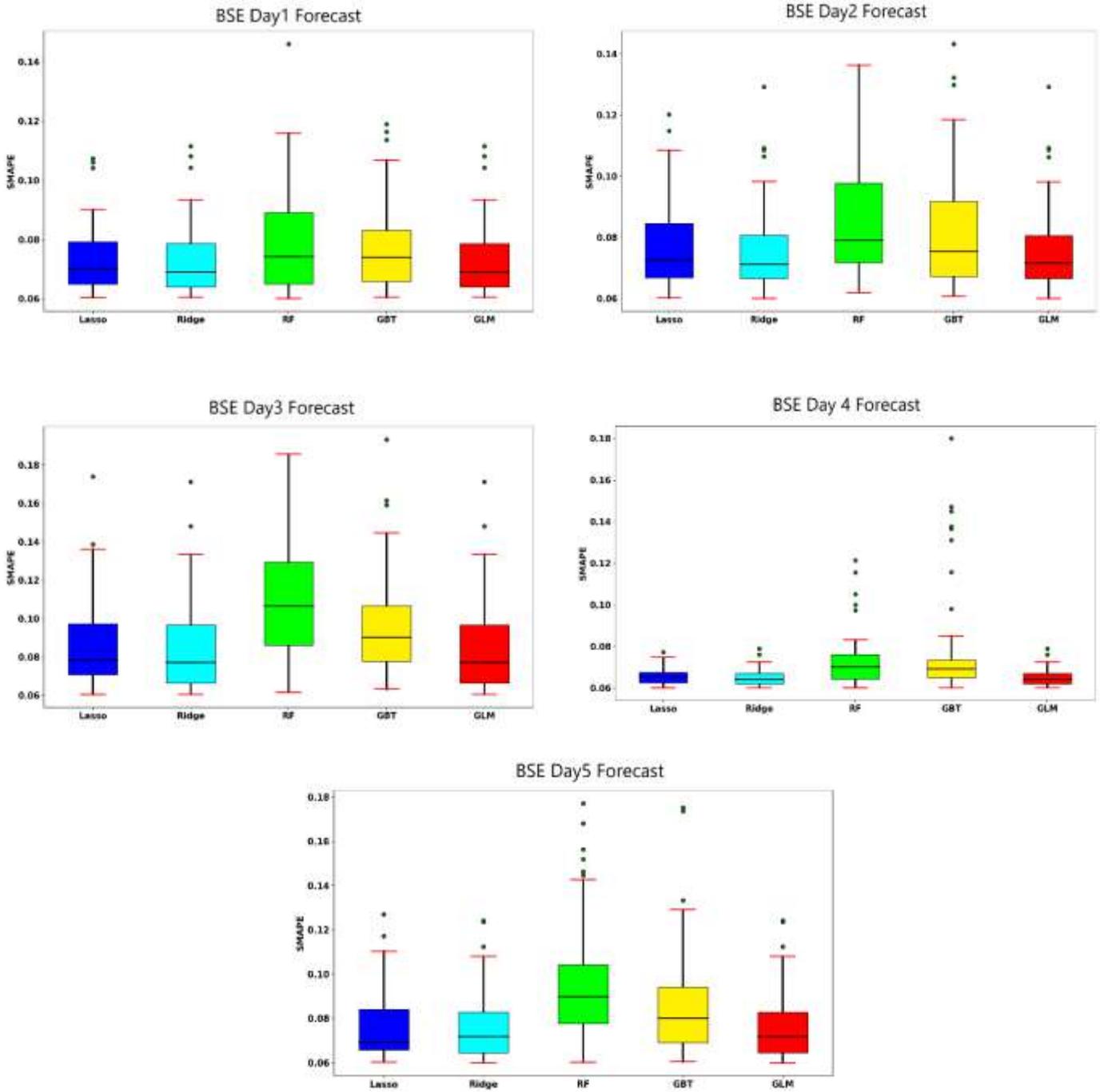

*Figure 6: Boxplot for BSE*

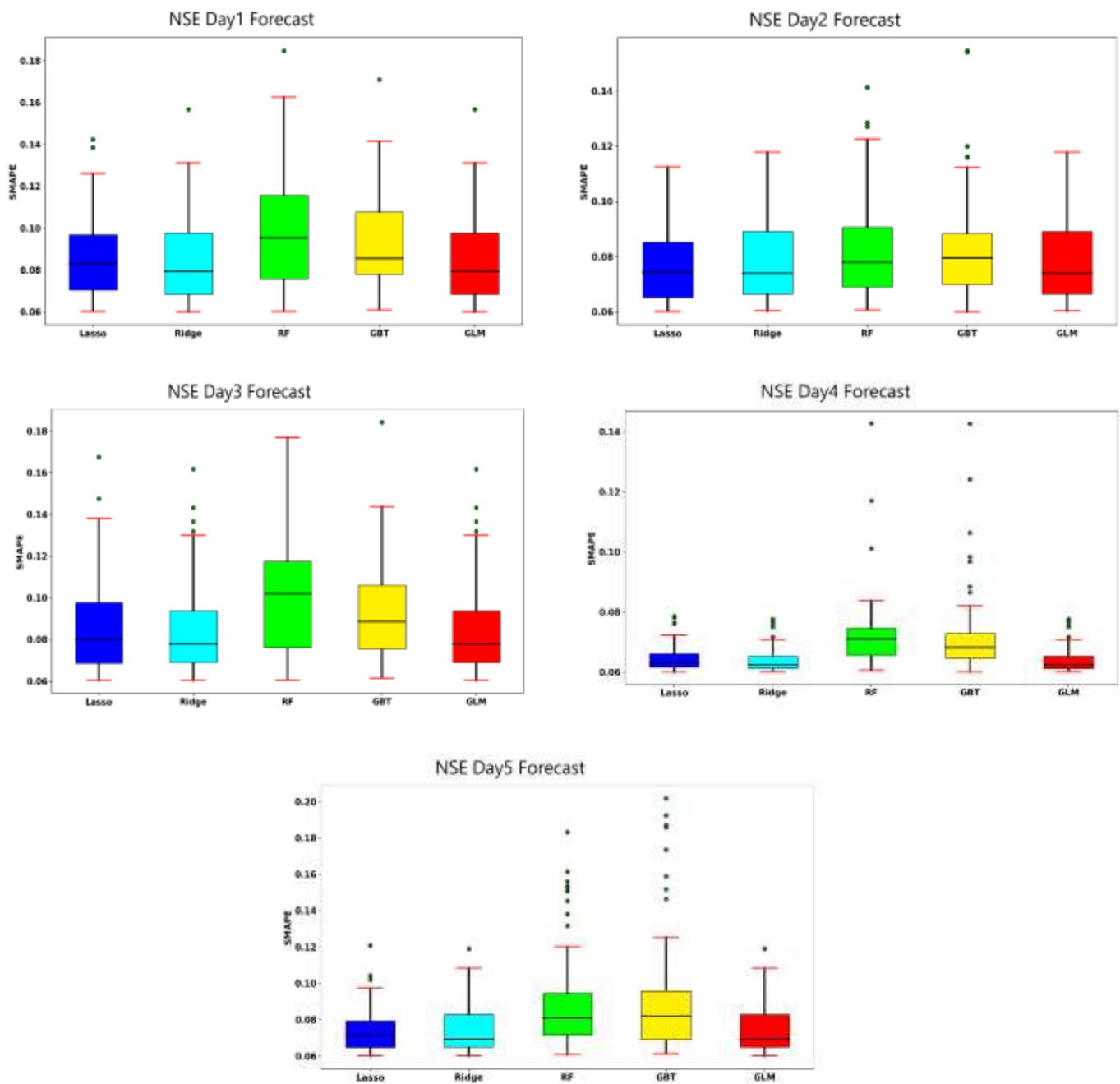

*Figure 7*: *Boxplot for NSE*

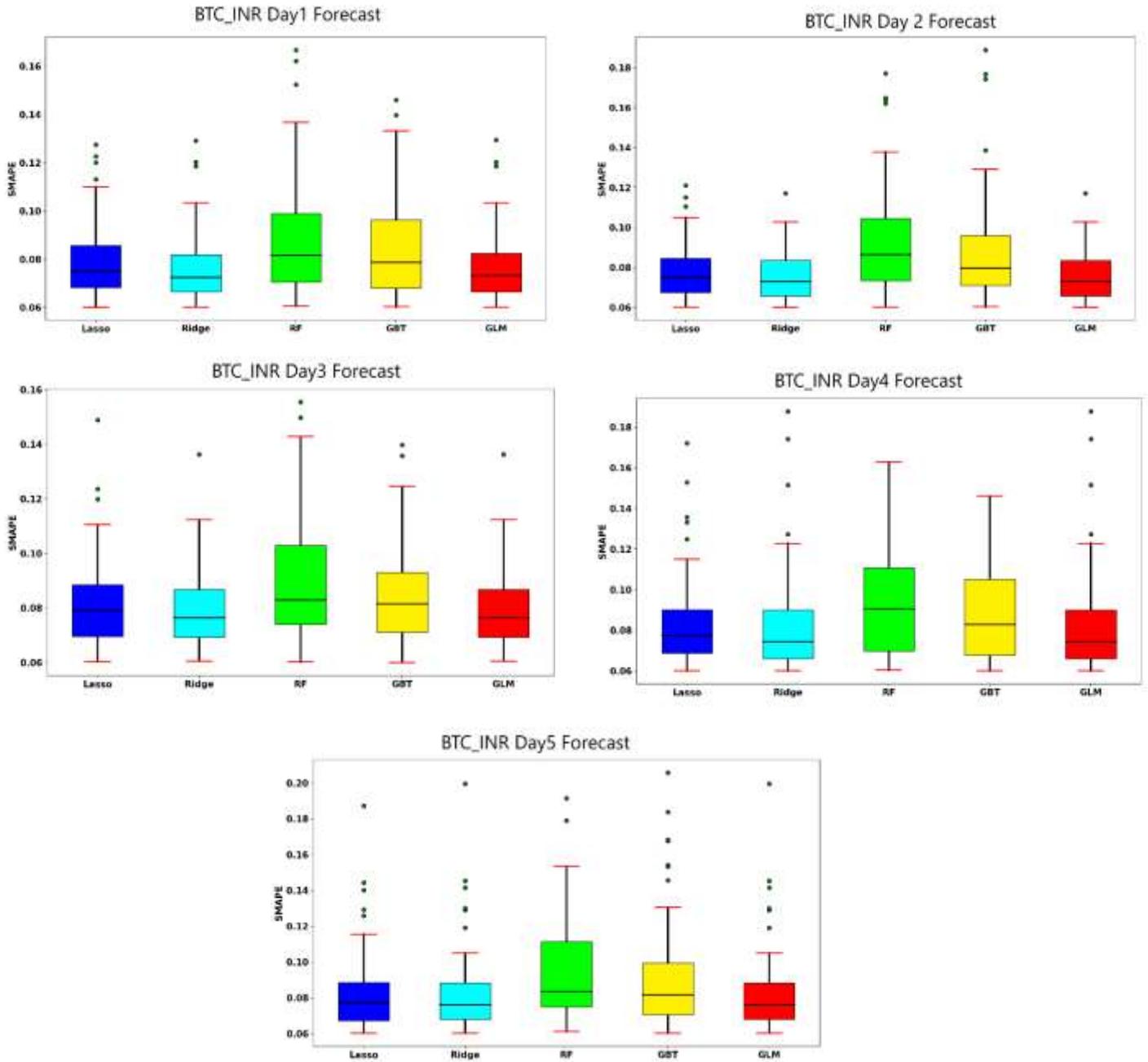

*Figure 8: Boxplot for BTC-INR*

Boxplots depicted in Figures 9, 10 and 11, shows the comparative result of Actual price vs forecasted price from all 5 models on all 3 datasets. These box plots also show how similar boxplot are with actual. In our experiment, Lasso, Ridge and GLM are proven to be most similar to actual price but RF and GBT fail to forecast accurately. In BTC-INR we can see boxplot having few outliers where it shows how many times the actual price varies from its own behaviour i.e. sudden rise or sudden fall. But an important thing to notice here is the models are able to forecast the sudden rise and sudden fall in the price. In this part again Lasso, Ridge, and GLM proved better than RF and GBT.

In some boxplots like NSE and BSE there is no outlier point. Then in these cases we can find similarity between Actual price and forecasted price using median and IQR of each boxplot.

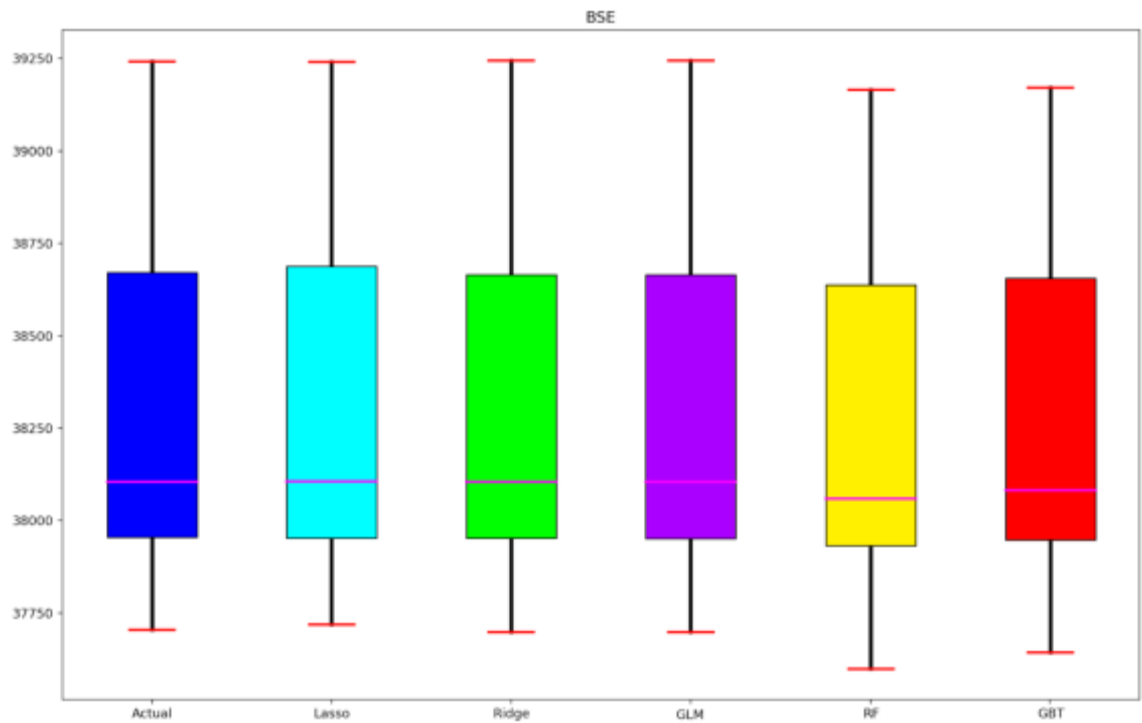

*Figure 9: Boxplot Actual Vs Model Forecast for BSE Data*

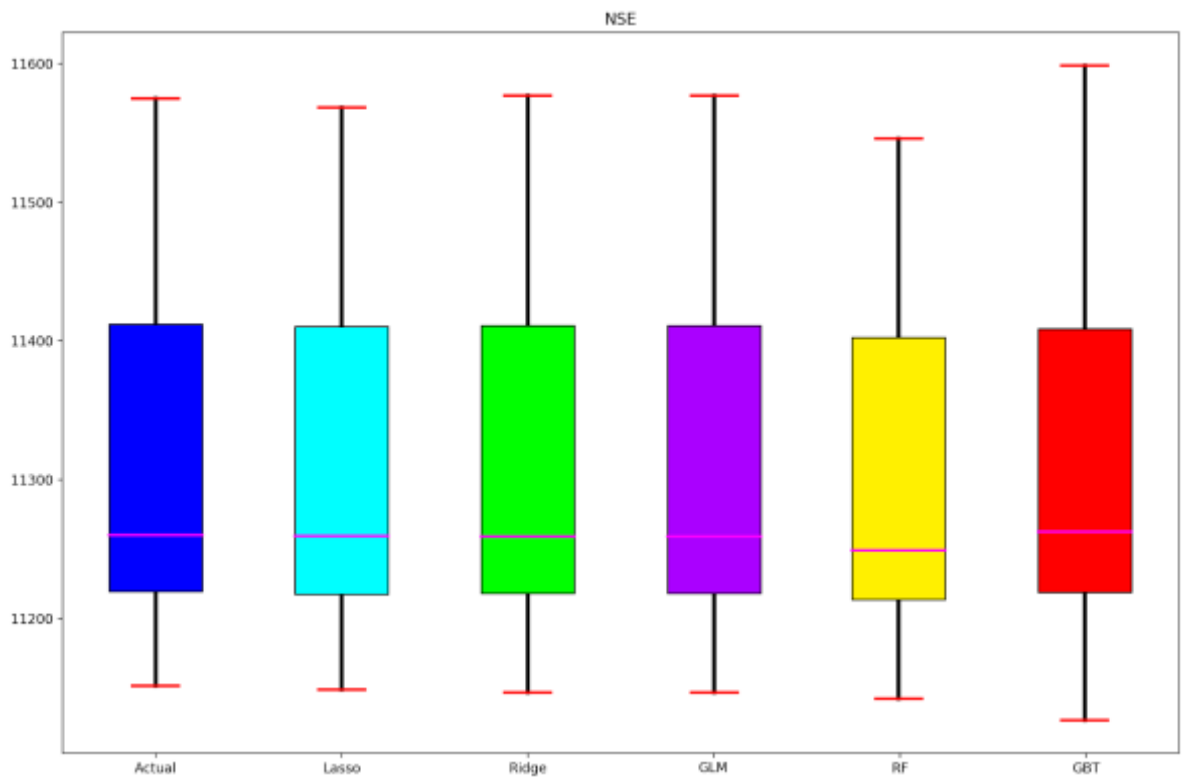

*Figure 10*: *Boxplot Actual Vs Model Forecast for NSE Data*

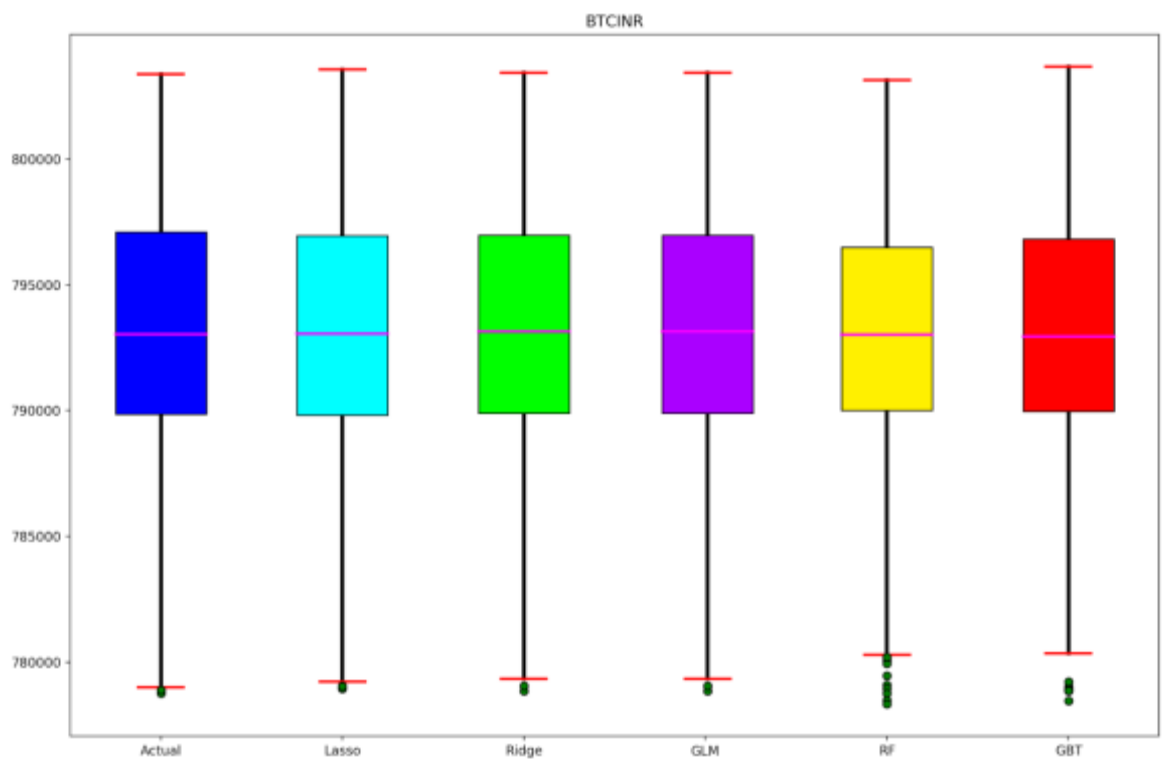

*Figure 11*: *Boxplot Actaul Vs Model Forecast for BTC-INR Data*

The hyperparameters combinations used for tuning all the models are presented in Table 7.

*Table 7: Hyper parameters for Grid Search*

| Model | Hyper parameters |
|---|---|
| Lasso | regParam = [0.01,0.02,0.03,0.04,0.05,0.1,0.2,0.3,0.4] elasticNetParam=0.0 |
| Ridge | regParam = [0.01,0.02,0.03,0.04,0.05,0.1,0.2,0.3,0.4] elasticNetParam=1.0 |
| Random Forest | maxDepth= [1,2,3] numTrees= [2,3,4] minInstancesPerNode= [1,2] |
| Gradient Boosting Tree | maxDepth= [1,2,3,4] maxBins= [4,8,16] minInstancesPerNode=[1,2] |
| Generalised Linear Model | regParam = [0.01,0.02,0.03,0.04,0.05,0.1,0.2,0.3,0.4] |

# Conclusions and Future Work

This work mainly focussed on developing an architecture for Real-time streaming and nowcasting the stock price with the help of Big Data Frameworks. In this work proposed a 2-stage methodology where chaos is modelled in the first stage followed by implemented sliding window approach for training on real-time data and then nowcasting the values with a frequency of every 5 minutes. From the results, it is clear that simple models such as Ridge Regression, Lasso Regression and GLM is performing better on small volume of data better than ensemble methods such as Random Forest and Gradient Boosting Trees. But if the data grows the complex algorithms like Random Forest and Gradient Boosting Trees might perform better.

In future, we can use Deep learning models for real-time nowcasting and also include other features for nowcasting like newsfeeds of the respective companies or markets, traded volumes. The same study can be also implemented on Macroeconomics time series data like Consumer price index (CPI).